\documentclass[10pt,twocolumn,letterpaper]{article}
\usepackage{iccv}
\usepackage{times}
\usepackage{epsfig}
\usepackage{graphicx}
\usepackage{amsmath}
\usepackage{amssymb}
\usepackage{csquotes}
\usepackage{algorithm}
\usepackage{algpseudocode}
\usepackage{booktabs}
\usepackage{array}
\usepackage{multirow}
\usepackage{xcolor}
\usepackage{enumitem}
\usepackage{subcaption}
\usepackage[accsupp]{axessibility}

\usepackage[pagebackref=true,breaklinks=true,letterpaper=true,colorlinks,bookmarks=false]{hyperref}

\iccvfinalcopy 

\def\iccvPaperID{10248} 

\newcommand{\DProbe}{\mathcal{D}_{probe}}

\ificcvfinal\fi

\begin{document}

\title{Corrupting Neuron Explanations of Deep Visual Features}

\author{
\textbf{Divyansh Srivastava}\\
UCSD CSE \\
{\tt\small ddivyansh@ucsd.edu}
\and
\textbf{Tuomas Oikarinen} \\
UCSD CSE \\
{\tt\small toikarinen@ucsd.edu}
\and
\textbf{Tsui-Wei Weng} \\
UCSD HDSI \\
{\tt\small lweng@ucsd.edu}
}

\maketitle
\ificcvfinal\thispagestyle{empty}\fi

\newcommand{\divyansh}[1]{\textcolor{blue}{#1}}

\begin{abstract}
    The inability of DNNs to explain their black-box behavior has led to a recent surge of explainability methods. However, there are growing concerns that these explainability methods are not robust and trustworthy. In this work, we perform the first robustness analysis of Neuron Explanation Methods under a unified pipeline and show that these explanations can be significantly corrupted by random noises and well-designed perturbations added to their probing data. We find that even adding small random noise with a standard deviation of 0.02 can already change the assigned concepts of up to 28\% neurons in the deeper layers. Furthermore, we devise a novel corruption algorithm and show that our algorithm can manipulate the explanation of more than $80\%$ neurons by poisoning less than 10\% of probing data. This raises the concern of trusting Neuron Explanation Methods in real-life safety and fairness critical applications. 
\end{abstract}

\begin{figure*}[!t]
  \centering
  \includegraphics[height=.43\linewidth]{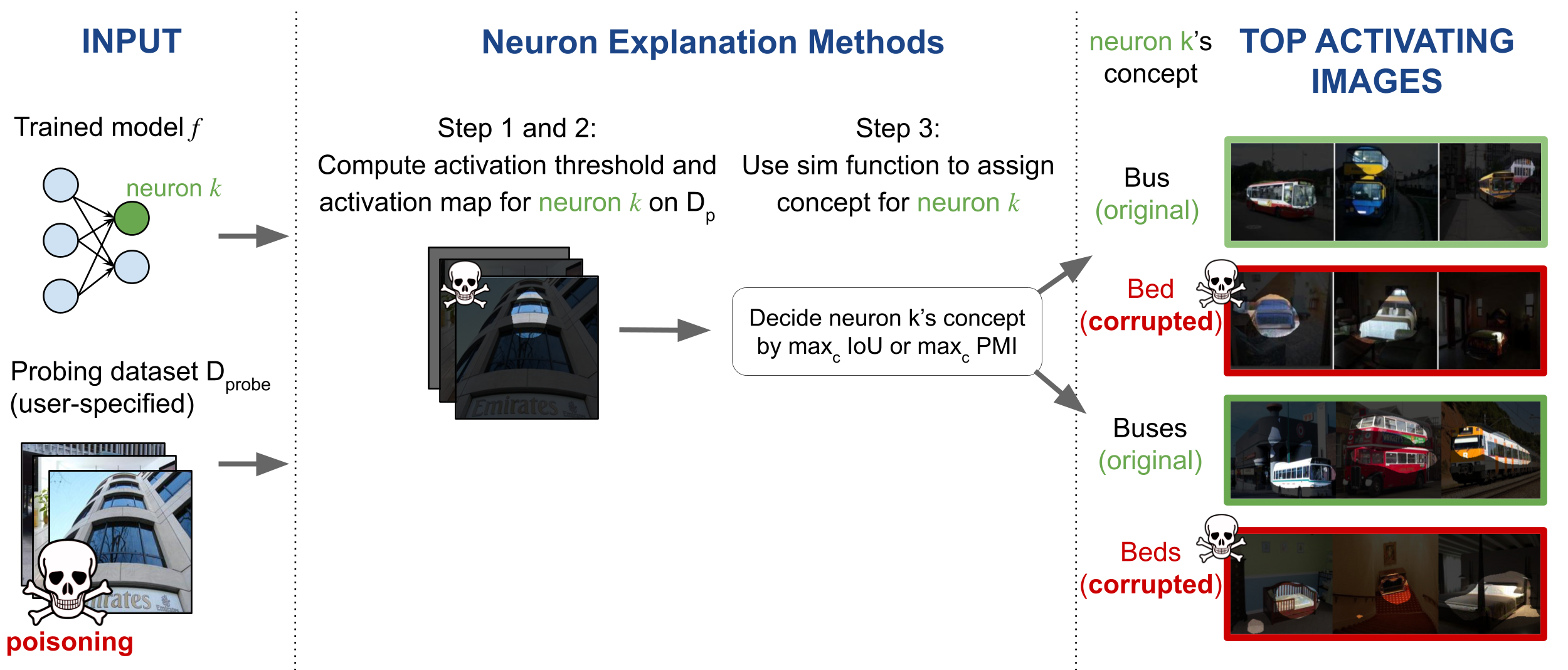}
  \caption{Pipeline depicting corruption of probing dataset to manipulate concepts of Neuron Explanation methods. \textcolor{red}{Red} words and outlines indicate parts of pipeline which have been manipulated due to corruption of the probing dataset.}
  \label{fig:Pipeline depicting attack on Neuron Interpretability methods}
\end{figure*}

\section{Introduction}
Deep neural networks (DNNs) have revolutionized many areas of Machine Learning and have been successfully used in various domains, including computer vision and natural language processing. However, one of the main challenges of DNNs is that they are black-box models with little insight into their inner workings. This lack of explainability can be problematic, particularly when using DNNs in safety and fairness critical real-world applications, such as autonomous driving, healthcare or loan/hiring decisions. Hence, it is imperative to explain the predictions of DNNs to increase human trust and safety of deployment. 

The need for explainability has led to a recent surge in developing methods to explain DNN predictions. In particular, \textit{Neuron Explanation methods} (NEMs) ~\cite{netdissect2017, bau2020understanding, mu2020compositional, hernandez2021natural} have attracted great research interest recently by providing global description of a DNN's behavior. This type of method explains the roles of each neuron in DNNs using human-interpretable concepts (e.g. natural language descriptions). Representative methods include Network dissection~\cite{netdissect2017, bau2020understanding}, Compositional Explanations ~\cite{mu2020compositional} and MILAN~\cite{hernandez2021natural}.

There has been recent interest in using NEMs to explain DNNs in safety critical tasks such as medical tasks, including brain tumor segmentation ~\cite{natekar2020demystifying} and medical image analysis ~\cite{salahuddin2022transparency}. However, despite of the excitement of NEMs, unfortunately the \textit{explanations} provided by NEMs could be significantly corrupted and not trustworthy, as first shown in this work. The untrustworthiness of explanation may result in negative consequences and misuse. For example, imagine an auditor uses a NEM to monitor potential biases in computer vision models by inspecting whether there exists neurons activating for specific skin colors. Such auditor might depend on user-provided probing data as it is impossible for auditors to collect validation datasets for every task, and would rely on manual inspections of images to find corruption. This would incentivize the developer of a biased model to fool NEMs by providing a corrupted probing dataset that bypasses manual inspection since perturbations are imperceptible by human eyes. This raises serious concerns about utilizing these NEMs for trusting model prediction.

In this paper, we would like to bring this awareness to the research community by performing the first robustness evaluation of Neuron Explanation methods.  Specifically, we unify existing NEMs under a generic pipeline and develop techniques to corrupt the neuron descriptions with a high success rate. We show that the neuron explanations could be significantly corrupted by both random noises and well-designed perturbation by only lightly corrupting the probing dataset in NEMs. We summarize our contributions in this work below:

\begin{enumerate}
    \item We are the first to present a unified pipeline for Neuron Explanation methods and show that the pipeline is prone to corruptions on the input probing dataset.
    \item We design a novel corruption on the probing dataset to manipulate the concept assigned to a neuron. Further, our formulation can manipulate the pipeline without explicit knowledge of the similarity function used by Neuron Explanation methods.
     \item We conduct the first large-scale study on the robustness of NEMs in terms of random noise and designed corruptions on the probing dataset. We show that even Gaussian random noise with a low standard deviation of 0.02 can manipulate up to 28\% of neurons, and our proposed algorithm can further corrupt the concepts of more than 80\% of neurons by poisoning less than 10\% probing images. This raises concerns about deploying NEMs in practical applications.
\end{enumerate}

\section{Related Work}
The mainstream interpretability methods to explain DNN's behavior can be categorized into two branches: one is at the instance level and the other is at the model/neuron level. The NEMs that we focus on in this paper provide explanations at the neuron/model level, and we provide a brief introduction on the instance-level interpretability method below. 

\vspace{-3mm}

\paragraph{Feature-attribution Explanations for DNNs.} Instance-level interpretability methods are also often known as feature-attribution methods, as they explain the model decision on each input instance by examining which input features (pixels) contribute the most towards a model’s output. This type of method includes CAM~\cite{zhou2016learning}, GradCAM~\cite{gradcam}, Integrated Gradients~\cite{sundararajan2017axiomatic}, SmoothGrad~\cite{smilkov2017smoothgrad}, and DeepLIFT~\cite{shrikumar2017learning}. Unfortunately, recent work has shown that the interpretations provided by instance-level interpretability methods can be easily manipulated~\cite{zheng2019analyzing,alvarez2018robustness,ghorbani2019interpretation}, which has increased concerns about utilizing these methods for trusting model prediction. It is worth noting that while the robustness of the \textit{Feature-attribution} methods is being investigated, the robustness of \textit{Neuron Explanation methods} has remained unexplored, which is thus the main focus of this work.  

\vspace{-3mm}

\paragraph{Neuron Explanation methods (NEMs) for DNNs.} 

NEMs explain model behavior on a global level by assigning descriptions to \textit{individual neuron}. Some of the representative methods in this class include Network Dissection~\cite{netdissect2017, bau2020understanding}, Compositional Explanations ~\cite{mu2020compositional}, and the recent work MILAN~\cite{hernandez2021natural}. Network Dissection~\cite{netdissect2017} is the first work proposing understanding DNNs by inspecting the functionality of each neuron with a set of pre-defined \textit{concepts}. The key idea of Network Dissection~\cite{netdissect2017} is to detect the concept of a neuron by matching the neuron activations to the pattern of a ground-truth concept label mask. It assigns an atomic concept to a neuron and does not capture relationships between different concepts. The work Compositional Explanations~\cite{mu2020compositional} builds on Network Dissection and generates explanations by searching for logical forms defined by a set of composition operators over atomic concepts. The work MILAN~\cite{hernandez2021natural} is a recent NEM that provides a natural language description of neurons. The idea of MILAN is to represent a neuron with a set of probing image that maximally activate it and obtain a description that maximizes pointwise mutual information of this set. Note that ~\cite{bau2020understanding} is an extension of ~\cite{netdissect2017} with the modification of obtaining pixel-level concepts from a segmentation model. This setting also falls into the general NEMs pipeline that we have described in Sec \ref{Unifying Neuron Interpretability methods}, and can be handled by our unified pipeline in Sec \ref{Problem Formulation}. In this work, we focus on Network Dissection~\cite{netdissect2017, bau2020understanding} and MILAN~\cite{hernandez2021natural}, which are representative methods in this class, but our formulation is general and applicable to all NEMs.

We highlight that this is the first work to study the sensitivity of neuron \textit{explanations} from NEMs, while prior work mostly focused on manipulating DNN’s \textit{predictions} and requires very different methods. Prior work on Cut-and-paste/copy-paste methods~\cite{casper2022robust, hernandez2021natural} aim to manipulates a DNN's \textit{prediction} by adding patch of unrelated object on input image at the test time using information from NEMs explanation. In contrast, this work shows that the NEMs themselves are vulnerable, and can produce wrong explanations when their inputs are corrupted. As NEMs have started being adopted to explain DNNs at neuron level in safety-critical domains like healthcare~\cite{natekar2020demystifying, salahuddin2022transparency}, it is critical to understand NEMs vulnerabilities and to caution the community against blindly trusting explanations from NEMs.

\section{Our Approach}
In this section, we describe our approach to evaluate the robustness of the \textit{Neuron Explanation Methods} (NEMs). In Sec \ref{Unifying Neuron Interpretability methods}, we show that existing NEMs can be unified into a generic pipeline which is crucial to study the robustness of this class of methods. In Section \ref{Problem Formulation}, we present a novel dataset corruption algorithm and show that it is possible to exploit the vulnerability in the probing step of the unified pipeline to manipulate neuron explanations.
\subsection{Unifying Neuron explanation pipelines}
\label{Unifying Neuron Interpretability methods}
In this section, we present a unified pipeline for NEMs and show that the representative methods, including Network Dissection~\cite{netdissect2017} and MILAN~\cite{hernandez2021natural}, are special instances of this generic pipeline. Define $X = \{x \;  |  \; x \in \mathbb{R}^{m \times n \times 3}\}$ as a set of images, $Y$ as a set of labels, and $f: X \rightarrow Y$ to be a Convolution Neural Network (CNN) with $L$ layers. Further, let $i$ represents a neuron unit in $f$, and $f_{i}(x)$ denote $i$-th neuron's activation map at input image $x \in \mathbb{R}^{m \times n \times 3}$, which is a three dimensional tensor. Network Dissection and MILAN describe individual channels in a convolutional layer and call them individual \textit{neuron}, because features in the same channel correspond to the same (filter) weight. In the remaining paper, we will use \textit{neuron}, \textit{unit} or \textit{channel} interchangeably to refer to the individual channels of a convolutional layer.

Let $\DProbe$ represent the probing dataset used for computing activations of a neuron, and $\mathcal{D}_{c}$ represent the set of concepts (i.e. explanations) that can be assigned to a neuron. The overall unified NEMs pipeline returns an explanation or concept $c_{i}^{*} \in \mathcal{D}_{c}$ for a neuron $i$ in the following three steps:
\begin{itemize}[leftmargin=*]
    \item \textbf{Step 1.} Compute activation threshold $T_{i}$ as the top $\eta$ quantile level for probability distribution $\mathbb{P}$ of activation maps $f_{i}(x), \; \forall x \in \DProbe$:
    \begin{equation} \label{Step 1: NIMs}
        T_{i} = \left[ \; \underset{t}{\max} \; \mathbb{P}_{x} \left[ \; f_{i}(x) > t \; \right] \geq \eta \; \right]
    \end{equation}
    \item \textbf{Step 2.} Create binary activation map $M_{i}(x) \in \{ 0, 1\}^{m \times n}$ by upsampling $f_{i}(x)$ to match the size of input image $x \in \mathbb{R}^{m \times n \times 3}$ using a biliner interpolation function $\mathcal{B}$, and thresholding with $T_{i}$:
    \begin{equation} \label{Step 2: NIMs}
        M_{i}(x) = \left[ \; \mathcal{B}(f_{i}(x)) \geq T_{i} \; \right]
    \end{equation}
    \item \textbf{Step 3.} Assign concept with the maximum value on the similarity function $\textsf{sim}$ to neuron $i$:
    \begin{equation} \label{Step 3: NIMs}
        c_{i}^{*} = \underset{c \in \mathcal{D}_{c}}{\textrm{argmax}} \; \textsf{sim}(\{(x, M_{i}(x))  \;| \; x \in \DProbe\}, \; c)
    \end{equation}
\end{itemize}

As we can see, step 1 and 2 are method-agnostic, as they only associate with the network $f(x)$ to be dissected and the probing dataset $\DProbe$, while step 3 will use a similarity function, $\textsf{sim}$, that is specific to different methods as discussed below.

\paragraph{(I) $\textsf{sim}$ of Network Dissection~\cite{netdissect2017, bau2020understanding}:} For Network Dissection, let $L_{c}(x) \in \{ 0, 1\}^{m \times n}$ represent the binary ground-truth segmentation mask for a concept $c \in \mathcal{D}_{c}$, which indicate whether each pixel in $x$ is associated with the concept $c$. In Network dissection, $L_{c}(x)$ can be obtained either by outsourcing experts to densely annotate probing images $x \in \DProbe$ with pixel-level concepts (e.g. broden in  \cite{netdissect2017}), or using a pre-trained segmentation model \cite{bau2020understanding} to perform pixel-level segmentation for each probing image in order to obtain the pixel-level concept information. The concept set $\mathcal{D}_{c}$ is a closed set containing all the possible explanations/concepts that can be assigned to a pixel. Further, Network Dissection uses $\eta = 0.005$ to calculate a neuron's activation thresholds. The similarity function $\textsf{sim}$ calculates \textit{Intersection Over Union} (IOU) for a concept $c$ and neuron $i$ by examining the overlap between $M_{i}(x)$ and the ground truth segmentation mask $L_{c}(x)$ over $\DProbe$ as
\begin{equation} \label{Step 3: Network Dissection}
        \begin{split}
        \textsf{sim}(\{(x, M_{i}(x)) \; | \; x \in \DProbe\}, \; c)\\ 
        = \frac{\sum_{x \in \DProbe} | M_{i}(x) \cap L_{c}(x)|}{\sum_{x \in \DProbe} | M_{i}(x) \cup L_{c}(x)|}
        \end{split}
\end{equation}

\paragraph{(II) $\textsf{sim}$ of MILAN~\cite{hernandez2021natural}:} MILAN provides a natural language description for a neuron, and $\mathcal{D}_{c}$ is an open set containing all possible natural language descriptions. The similarity function $\textsf{sim}$ can be split into two steps. First, we calculate an exemplar set $E_{i}$ containing $k$ top-activating images $x \in \DProbe$ on a neuron $i$ as given Eq \ref{Step 1: MILAN}, and then calculate the pointwise mutual information (PMI) for concept $c$ and exemplar set $E_{i}$ as given in Eq \ref{Step 2: MILAN}.
\begin{subequations}
    \begin{equation} \label{Step 1: MILAN}
        E_{i} = \{(x,M_{i}(x)) | \; M_{i}(x) = \left[ \; \mathcal{B}(f_{i}(x)) \geq T_{i} \; \right] \}
    \end{equation}
    \begin{equation} \label{Step 2: MILAN}
        \begin{split}
            \textsf{sim}(\{(x, M_{i}(x)) \; | \; x \in \DProbe\}, \; c) \\
            = \textrm{log} \: p(c \: | \: E_{i}) - \textrm{log} \: p(c)
        \end{split}
    \end{equation}
\end{subequations}
where $p(c)$ is the probability that a human would use the description $c$ for any neuron, and $p(c|E)$ is a distribution over image captions. In MILAN, these two terms are learned and approximated by pre-trained neural networks. Particularly, $p(c)$ is approximated by 
a two-layer LSTM network trained on the text of MILANNOTATION, and $p(c|E)$ is approximated by a modified Show-Attend-Tell image description model trained on MILANNOTATIONS dataset. Implementation details can be found in the Appendix of MILAN~\cite{hernandez2021natural}.
\subsection{Formulation}
\label{Problem Formulation}
As shown in the unified pipeline in Section \ref{Unifying Neuron Interpretability methods}, Neuron Explanation methods require probing datasets $\DProbe$ to compute the activation map and threshold for each neuron in Step 1 and the similarity measure in Step 3. The probing dataset is domain-specific and generally collected by end users working with DNNs. This makes it vulnerable to noises and adversaries who can corrupt the probing dataset with the intent to manipulate concepts generated by NEMs. The data poisoning can easily be carried out in settings involving an exchange of the probing dataset over networks or by an adversary with backdoor access to the probing dataset. Until this work, it is not known if there indeed exists corruption that could fool NEMs. Following this discussion, we focus on analyzing the robustness of  NEMs against random noise as well as crafted perturbations, where we design a novel data corruption algorithm on the probing dataset that can manipulate NEMs explanations with a high success rate.

\begin{figure}[h]
 \centering
\begin{subfigure}{\linewidth}
  \centering
  \includegraphics[width=\linewidth]{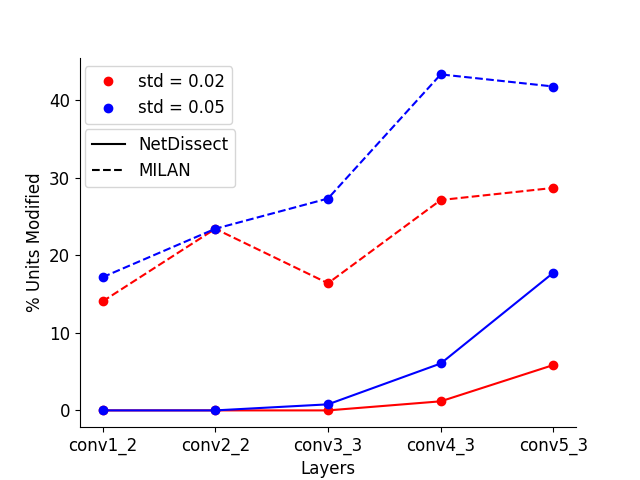}
  \caption{}
  \label{fig:Gaussian Noise/Network Dissection/Effect of random noise on layer interpretability by count}
\end{subfigure}
\begin{subfigure}{\linewidth}
  \centering
  \includegraphics[width=\linewidth]{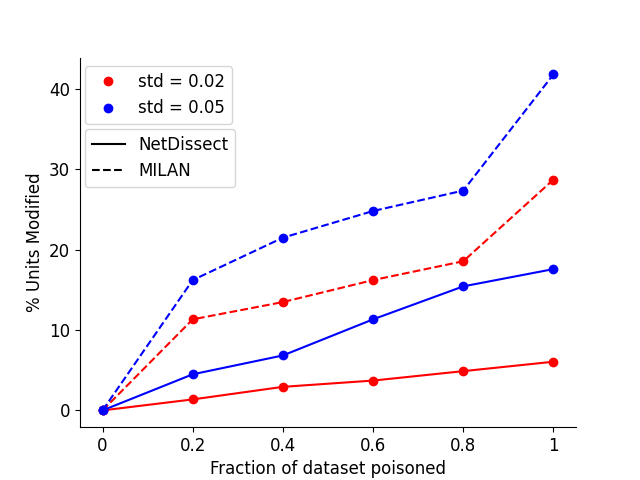}
  \caption{}
  \label{fig:Gaussian Noise/Network Dissection/Effect of random noise on concept interpretability in layer conv5_3}
\end{subfigure}
\centering
\caption{Effect of Gaussian random noise on neuron explanations for VGG16-Places365. The red and blue lines visualize results for std = 0.02 and std = 0.05, respectively. Fig (a) visualizes the percentage of manipulated neurons explanation (y-axis) in layers of VGG16-Places365 (x-axis) for noise std of 0.02 and 0.05. Fig (b) visualizes the percentage of manipulated neurons explanation (y-axis) with increasing fraction of poisoned images (x-axis) for noise std of 0.02 and 0.05 in conv5\_3.}
\label{fig:Effect of random noise on interpretability}
\end{figure}

\paragraph{(I) Corruption by Random noise.} To evaluate the effect of random noise to the images in the probing dataset, we define $\mathcal{D'}_{probe} = \{ x + \mathcal{N}(0, \sigma^2), \forall x \in \DProbe \}$ and run NEM with the corrupted probing dataset $\mathcal{D'}_{probe}$. This simple corruption method can manipulate concepts of upto 28\% neuron with a low std of 0.02 as shown in Fig \ref{fig:Effect of random noise on interpretability}.

\paragraph{(II) Corruptions by designed perturbations.} Given a neuron $i$, our goal is to compute the minimum corruption such that the corresponding concept $c_{i}^{*}$ is changed:
\begin{equation} \label{eq:min_adv_pert}
    \begin{aligned}
        & \underset{\delta_{j}}{\min} \, && \underset{j}{\textrm{max}} \;\; ||\delta_j||_{\infty}  \\
        & \textrm{s.t.} && c_{i}^{*}(\{x_{j}\}) \neq c_{i}^{*}(\{x_{j}+\delta_j\}) \\
        &&& x_j+\delta_j \in [0,1]^l, \, \forall x_j \in \DProbe.
    \end{aligned}
\end{equation}

This optimization problem is a general non-convex function as deep neural network $f$ is generally non-convex, and so are its activation maps. In order to optimize it while considering the constraint with $c_{i}^{*}$, we have devised a differentiable objective function that is more amenable. To start with, recall that Eq \ref{Step 3: NIMs} measures the similarity between a neuron's activation map and the concepts, and assigns the concept with the highest similarity score to the neuron. We can thus manipulate the neuron's concept by designing corruptions for each image in the probing dataset that can change the neuron's activation map. Specifically, we define $act_{i,c}^{avg}(x)$ as the average of activation values of pixels that are associated with concept $c$ and image $x \in \DProbe$ in the receptive field of neuron $i$. Formally, we define
\begin{equation} \label{adversarial objective function}
    act_{i,c}^{avg}(x) = (\frac{\sum_{k} \mathcal{B}(f_{i}(x))_{k} \times L_{c}(x)_{k}} {\sum_{k} L_{c}(x)_{k}})
\end{equation}
where $k$ represents a location in the 2-D map $\mathcal{B}(f_{i}(x))$ and $L_{c}(x)$. We overload the definition of binary segmentation mask for an image $x$ and concept $c$, $L_{c}(x)$, defined in the context of Network Dissection to all the NEMs. Simply, $L_{c}(x)$ is $1$ at pixels that contain the concept $c$ in image $x$ and $0$ elsewhere. $L_{c}(x)$ can be obtained in two ways:
\begin{itemize}
    \item $L_{c}(x)$ from ground truth segmentations: This is the simplest of two scenarios where per-pixel segmentation data can be directly obtained. This is the typical setting for Network Dissection, where the ground truth segmentation is available to the concept assignment function $\textsf{sim}$ either from the Broden dataset or a pre-trained segmentation model.
    \item $L_{c}(x)$ for general NEMs: Ground truth segmentation data might not always be available, for example, when segmentation data is not required for computing $\textsf{sim}$ as in MILAN, or the concept assignment function $\textsf{sim}$ including per-pixel segmentation data is not accessible to users. In this scenario, we can obtain a good approximation for $L_{c}(x)$ from the standard (non-corrupted) runs. Specifically, consider $\{x, M(x)\}$ pair that resulted in the assignment of label $c_{i}^{*}$ for a neuron $i$. We define $L_{c_{i}^{*}}(x)$ as
    \begin{equation}
        L_{c_{i}^{*}}(x) = M(x)
    \end{equation}
    This technique of defining $L_{c_{i}^{*}}(x)$ can be extended to all the NEMs if the ground truth segmentation data is unavailable. However, this reduces the scope of concepts that are available for manipulation of neurons in $f$, i.e., $\{ c_{i}^{*} \; | \; \forall i \in f \}$.
\end{itemize}

Using $act_{i,c}^{avg}(x)$, we define our objective function for neuron $i$ with concept $c_{i}^{*}$, probing image $x_j \in \DProbe$ and the target (manipulated) concept $t$ as
\begin{equation} \label{Attack: Targeted with minimum distance}
    \begin{aligned}
        & \underset{\delta_{j}}{\min} && act_{i,c_{i}^{*}}^{avg}(x_j+\delta_{j}) - act_{i,t}^{avg}(x_j+\delta_{j})  \\
        & \textrm{s.t.} && ||\delta_{j}||_{\infty} \leq \epsilon \\
        &&& x_j+\delta_{j} \in [0,1]^l
    \end{aligned}
\end{equation}

Therefore, instead of optimization Eq \ref{eq:min_adv_pert} directly, we use the designed surrogate function and solve Eq \ref{adversarial objective function}. The intuition is to decrease the activations of pixels that contain the source concept and increase the activation of pixels that contain the target concept for neuron $i$. Consequently, we can solve for the perturbation $\delta_{j}$ to poison each probing image $x_j$ nicely by solving Eq \ref{Attack: Targeted with minimum distance} with projected gradient descent~\cite{goodfellow2014explaining, madry2018towards}. 

\section{Experiment}
\begin{figure*}[ht]
    \captionsetup[subfigure]{labelformat=empty}
  \centering
  \begin{subfigure}[b]{.49\linewidth}
        \centering
        \includegraphics[width=\linewidth]{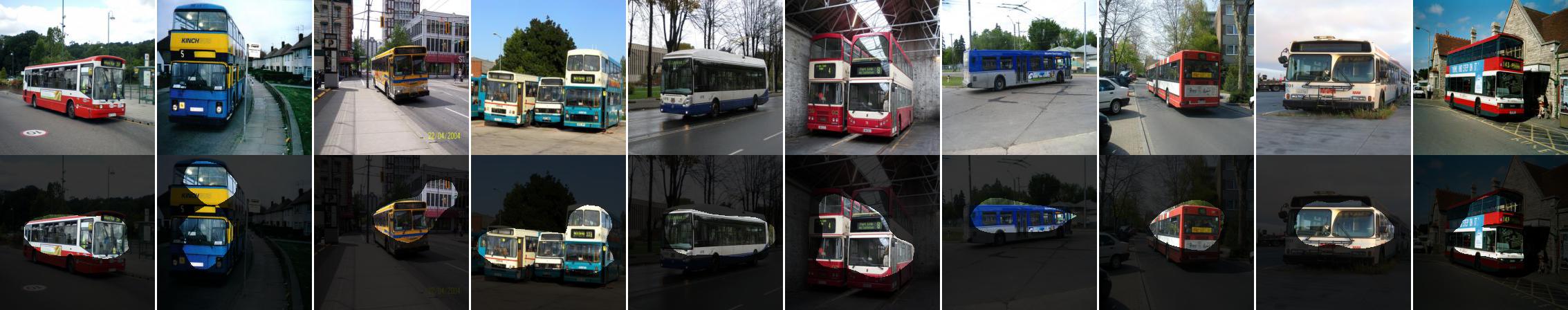}
        \caption{Original, Concept: bus, IOU: 0.15}
        \vspace{4pt}
    \end{subfigure}
    \hfill
    \begin{subfigure}[b]{.49\linewidth}
        \centering
        \includegraphics[width=\linewidth]{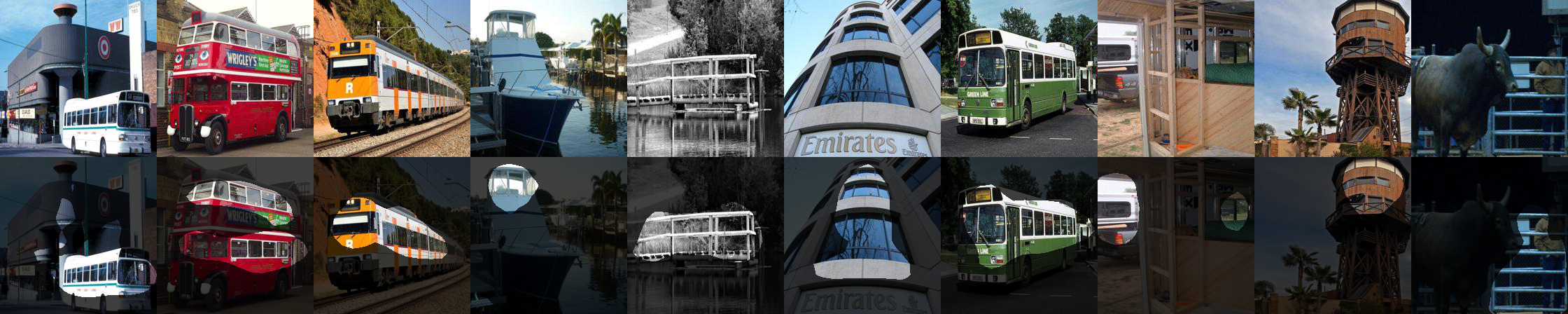}
        \caption{Original, Description: Buses}
        \vspace{4pt}
    \end{subfigure}    
    \hfill
    \begin{subfigure}[b]{.49\linewidth}
        \centering
        \includegraphics[width=\linewidth]{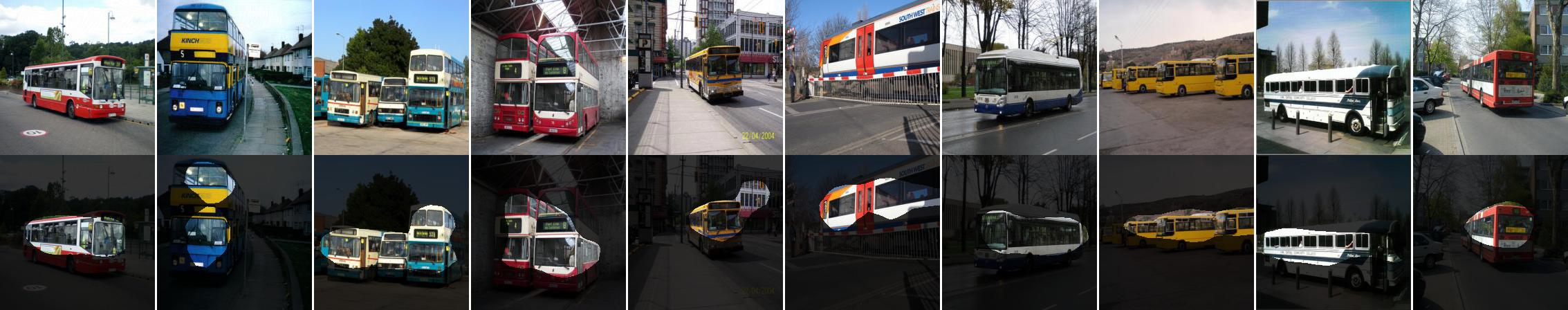}
        \caption{$\epsilon = 2/255$, {Concept: bus, IOU: 0.10}}
        \vspace{4pt}
    \end{subfigure}
    \hfill
    \begin{subfigure}[b]{.49\linewidth}
        \centering
        \includegraphics[width=\linewidth]{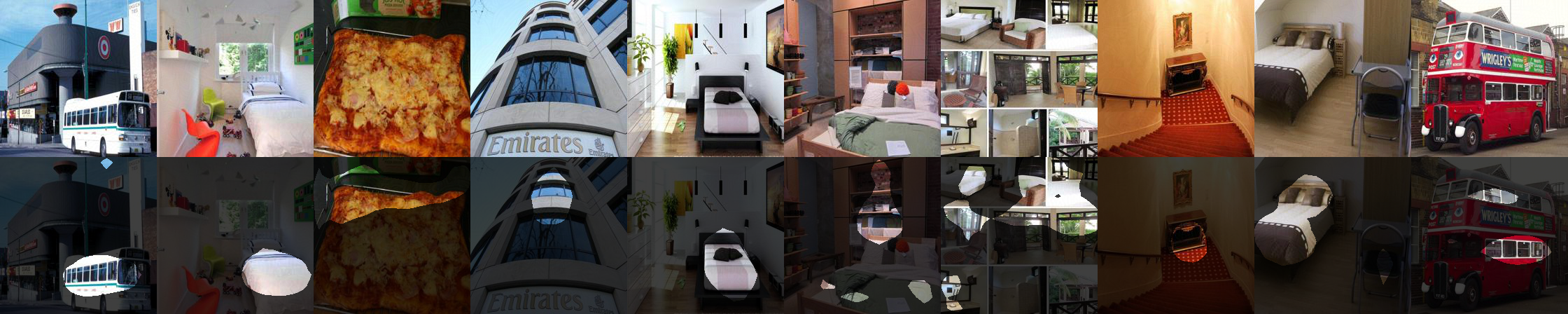}
        \caption{$\epsilon = 2/255$, \textcolor{red}{(Corrupted) Description: Bed frame}
        }
        \vspace{4pt}
    \end{subfigure}
    \hfill
    \begin{subfigure}[b]{.49\linewidth}
        \centering
        \includegraphics[width=\linewidth]{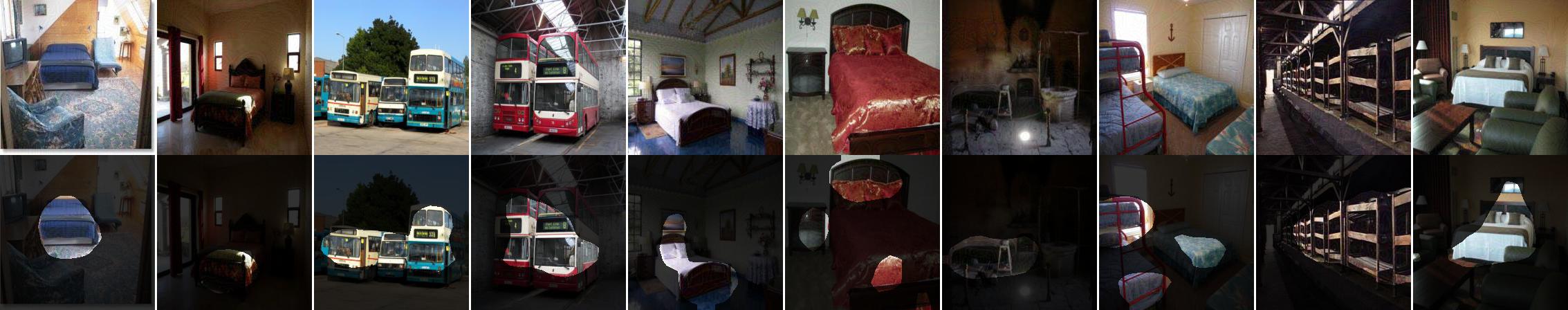}
        \captionsetup{justification=centering}
        \caption{$\epsilon = 3/255$, \textcolor{red}{(Corrupted) Concept: bed, IOU: 0.17}}
        \vspace{4pt}
    \end{subfigure}
    \hfill
    \begin{subfigure}[b]{.49\linewidth}
        \centering
        \includegraphics[width=\linewidth]{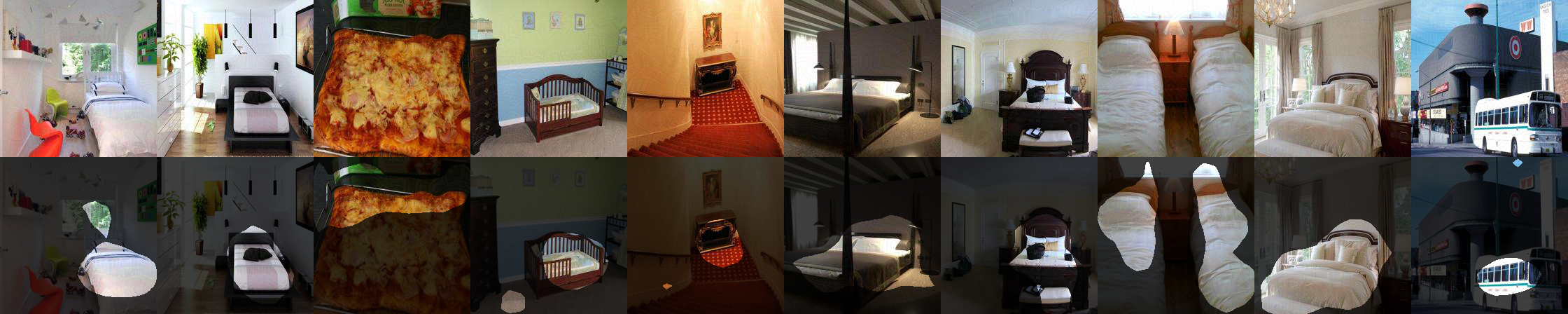}
        \caption{$\epsilon = 3/255$, \textcolor{red}{(Corrupted) Description: Beds}}
        \vspace{4pt}
    \end{subfigure}
    \caption{Targeted data corruption on Network Dissection (left) and MILAN (right) in VGG16-Places365 conv5\_3 layer to change concept of unit 191 from \textit{Buses} to \textit{Beds}. First row in subfigure visualizes top activating images from $\mathcal{D}_{probe}$ for neuron 191. Second row in each subfigure highlights pixels with value greater than activation threshold $T_{191}$.} 
    \label{fig:Visualize attack Unit 191 MILAN}
\end{figure*}
\begin{table}[t]
  \centering
  \begin{tabular*}{\linewidth}{l@{\extracolsep{\fill}}  r@{\extracolsep{\fill}} r@{\extracolsep{\fill}} r@{\extracolsep{\fill}} r@{\extracolsep{\fill}} r@{\extracolsep{\fill}} r@{\extracolsep{\fill}}}
  \toprule
    & & \multicolumn{5}{c}{Noise}\\
           \cmidrule(lr){3-7}
    Layer & & 0.01 & 0.02 & 0.03 & 0.04 & 0.05\\
    \midrule
    \multirow{2}{*}{conv1\_2} & mean & 0.936 & 0.892 & 0.884 & 0.870 & 0.843 \\
    & std & 0.172 & 0.208 & 0.222 & 0.196 & 0.203 \\
    \midrule
    \multirow{2}{*}{conv2\_2} & mean & 0.918 & 0.862 & 0.856 & 0.829 & 0.848 \\
    & std & 0.169 & 0.202 & 0.212 & 0.214 & 0.206 \\
    \midrule
    \multirow{2}{*}{conv3\_3} & mean & 0.922 & 0.886 & 0.845 & 0.813 & 0.813 \\
    & std & 0.181 & 0.214 & 0.226 & 0.235 & 0.238 \\
    \midrule
    \multirow{2}{*}{conv4\_3} & mean & 0.896 & 0.834 & 0.778 & 0.765 & 0.735 \\
    & std & 0.207 & 0.238 & 0.259 & 0.259 & 0.263 \\
    \midrule
    \multirow{2}{*}{conv5\_3} & mean & 0.886 & 0.832 & 0.787 & 0.767 & 0.742 \\
    & std & 0.223 & 0.251 & 0.269 & 0.269 & 0.279 \\
    \bottomrule
    \end{tabular*}
  \caption{Average F1-BERT score between clean and manipulated descriptions after random noise data corruption. F1-BERT score between similar strings has its maximum value 1.0, and lower F1-BERT score indicates higher dissimilarity and stronger manipulation.}
  \label{tab:random noise BERT score}
\end{table}
In this section, we perform the robustness analysis of Neuron Explanation Methods (NEMs) and show that they are not robust against various corruptions of probing dataset, from small random noises to well-designed perturbations which are imperceptible to human eyes. We give an overview of our experimental setup in Section \ref{Experiment-Setup}, examine the effect of Gaussian random noise on neuron explanations in Section \ref{Experiment-Non adversarial attack}, present qualitative and quantitative analysis of our proposed corruption algorithm in Section \ref{Experiment-Adversarial attack}, and show that our designed data corruption algorithm manipulates neuron explanations with a significantly higher success rate compared to random noise in Section \ref{Comparing adversarial and non-adversarial attacks}. Note that we use the term \textit{data poisoning} and \textit{data corruption} in this section to refer to the corruption of probing dataset $\DProbe$.
\subsection{Overview}
\label{Experiment-Setup}
\paragraph{Baselines.}
As described earlier, there are mainly two approaches of NEMs (Network dissection and MILAN) and we have unified them into the same pipeline in section \ref{Unifying Neuron Interpretability methods}. Therefore, in the following, we will show the result of both standard (non-corrupted) and corrupted neuron explanations for both methods as a comprehensive study. 

\paragraph{Setup.} Our experiments run on a server with 16 CPU cores, 16 GB RAM, and 1 Nvidia 2080Ti GPU. Our codebase builds on the open source implementation of \href{https://github.com/CSAILVision/NetDissect-Lite}{Network Dissection}~\cite{netdissect2017} and \href{https://github.com/CSAILVision/NetDissect-Lite}{MILAN}~\cite{hernandez2021natural} released from their papers. We use pretrained \href{https://dissect.csail.mit.edu/models/vgg16_places365-0bafbc55.pth}{VGG16} on Places365\cite{zhou2017places} and pretrained Resnet50 on Imagenet, referred to as \textit{VGG16-Places365} and \textit{Resnet50-Imagenet} respectively. The choice of these models follows the setting in the Network Dissection and MILAN paper for fair comparison. We modify projected gradient descent implementation available in Torchattacks\cite{kim2020torchattacks} library for optimizing our objective function in Eq \ref{Attack: Targeted with minimum distance}.

\paragraph{Probing dataset.} NEMs require probing dataset to compute the activation threshold for neurons and generate an activation map, which is then used by the similarity function to assign concepts. In general, concept information does not need to be present in the probing dataset. For example, PNAS version of Network Dissection~\cite{bau2020understanding} uses validation dataset for probing the network and obtains per-pixel concept data from a pre-trained segmentation model. Similarly, MILAN uses held-out validation dataset of the trained model for probing DNNs, and uses a pre-trained model that generates descriptions for top activating maps. These are in contrast to the first version of Network Dissection ~\cite{netdissect2017} that use Broden dataset for probing DNNs and contains concept data. More precisely, Broden dataset contains pixel-level segmentation of concepts from six categories: (i) object, (ii) scenes,  (iii) object parts, (iv) texture, (v) materials, and (vi) color in a variety of contexts. In this section, we will run our experiments on two NEMs, Network Dissection using Broden dataset and MILAN to cover both the ends of this spectrum.

\paragraph{Evaluation metric.} The concept generated by Network Dissection is a term (e.g., ``bus") whereas MILAN generates a natural language description for each neuron unit (e.g., ``colorful dots and lines"). For Network Dissection, we define a neuron-unit as \textit{manipulated} if the concept assigned by Network Dissection is changed after probing dataset is corrupted. For MILAN, we use F1-BERT score~\cite{ghorbani2019interpretation} to quantify the change in description between the clean and manipulated descriptions. Further, to compare MILAN and Network Dissection, we define a neuron unit as ``manipulated" for MILAN if the F1-BERT score between clean description and manipulated description of a neuron $\leq 0.642$. Our estimate of this threshold is obtained by randomly selecting 500 clean and manipulated description pairs, manually evaluating whether the two descriptions have the same semantic meaning, and choosing a F1-BERT score cutoff where the number of false positives is equal to the number of false negatives. We have shown few examples in Appendix Fig A.4. We also report the F1-BERT scores alongside the manipulation success rate in the experiments to provide a complete picture. 


\begin{table*}[t]
\centering
\begin{tabular*}{.49\linewidth}{l@{\extracolsep{\fill}} r@{\extracolsep{\fill}} r@{\extracolsep{\fill}} r@{\extracolsep{\fill}} r@{\extracolsep{\fill}} r@{\extracolsep{\fill}}
r@{\extracolsep{\fill}}}
\toprule
& & \multicolumn{3}{c}{\textbf{VGG16-Places365}} \\
& \multicolumn{3}{c}{Untargeted} & \multicolumn{3}{c}{Targeted} \\
\cmidrule(lr){2-4}                       \cmidrule(lr){5-7}
Layer & $\epsilon=\frac{2}{255}$ & $\frac{4}{255}$ &  $\frac{6}{255}$ & $\frac{2}{255}$ & $\frac{4}{255}$ & $\frac{6}{255}$ \\
\midrule
conv1\_2 & 47.73 & 52.27 & 54.55 & 2.27 & 4.55 & 4.55 \\
\midrule
conv3\_3 & 34.88 & 58.14 & 67.44 & 4.65 & 13.95 & 25.58 \\
\midrule
conv4\_3 & 57.45 & 95.74 & 100.0 & 0.0 & 29.79 & 46.81 \\
\midrule
conv5\_3 & 70.59 & 96.08 & 96.08 & 7.84 & 49.02 & 56.86 \\
\bottomrule
\end{tabular*}
\begin{tabular*}{.49\linewidth}{l@{\extracolsep{\fill}} r@{\extracolsep{\fill}} r@{\extracolsep{\fill}} r@{\extracolsep{\fill}} r@{\extracolsep{\fill}} r@{\extracolsep{\fill}}
  r@{\extracolsep{\fill}}}
  \toprule
& & \multicolumn{3}{c}{\textbf{Resnet50-Imagenet}} \\
& \multicolumn{3}{c}{Untargeted} & \multicolumn{3}{c}{Targeted} \\
\cmidrule(lr){2-4}                       \cmidrule(lr){5-7}
Layer & $\epsilon =\frac{2}{255}$ & $\frac{4}{255}$ &  $\frac{6}{255}$ & $\frac{2}{255}$ & $\frac{4}{255}$ & $\frac{6}{255}$ \\
\midrule
layer1 & 16.25 & 76.25 & 87.5 & 9.68 & 20.97 & 32.26 \\
\midrule
layer2 & 29.17 & 85.42 & 91.67 & 1.90 & 20.00 & 42.86 \\
\midrule
layer3 & 75.34 & 98.63 & 98.63 & 3.90 & 27.63 & 46.05 \\
\midrule
layer4 & 79.35 & 98.63 & 98.63 & 2.70 & 22.97 & 32.43 \\
\bottomrule
\end{tabular*}
\caption{Percentage of units manipulated (higher means our corruption technique has stronger effects) in Network Dissection for VGG16-Places365(left) and Resnet50-Imagenet(right). More than 80\% neurons can be manipulated with untargeted data corruption of less than 10\% images in higher layers.}
\label{tab:units modified by attack on Network Dissection for VGG16-Places365}
\end{table*}

\begin{table}[t]
  \centering
  \begin{tabular*}{\linewidth}{l@{\extracolsep{\fill}}  r@{\extracolsep{\fill}} r@{\extracolsep{\fill}} r@{\extracolsep{\fill}}}
  \toprule
Layer & $\epsilon = \frac{2}{255}$ & $\epsilon = \frac{4}{255}$ &  $\epsilon = \frac{6}{255}$\\
\midrule
layer1  & 0.73 $\pm$ 0.29 &  0.63 $\pm$ 0.30 & 0.55 $\pm$ 0.32\\
\midrule
layer2  & 0.67 $\pm$ 0.22 & 0.56 $\pm$ 0.16 & 0.54 $\pm$ 0.20\\
\midrule
layer3  & 0.46 $\pm$ 0.22 & 0.50 $\pm$ 0.25 & 0.45 $\pm$ 0.21\\
\midrule
layer4  & 0.41 $\pm$ 0.29 & 0.42 $\pm$ 0.29 & 0.44 $\pm$ 0.31\\
\bottomrule
\end{tabular*}
\caption{Average F1-BERT score (lower means our corruption technique has stronger effects) and standard deviation for untargeted data corruption on MILAN for \textbf{Resnet50-Imagenet}. F1-BERT score between similar strings has its maximum value 1.0, and lower F1-BERT score indicates higher dissimilarity and more successful untargeted data corruption.}
\label{tab:Average BERT score and std of MILAN for untargeted attack on Resnet50-Imagenet}
\end{table}
\subsection{Random noise corruption of probing dataset}
\label{Experiment-Non adversarial attack}
In this section, we ask the following question: \textit{How does the concept of a neuron unit change when probing dataset is corrupted with random noise?}

Fig \ref{fig:Gaussian Noise/Network Dissection/Effect of random noise on layer interpretability by count} visualizes the percentage of neurons manipulated with Gaussian random noise in VGG16-Places365 for Network Dissection and MILAN. We observe that the higher layers of the network are more susceptible to being manipulated by Gaussian random noise. A noise with a low standard deviation of 0.05 can manipulate more than 17\% neurons for Network Dissection and over 40\% of neuron descriptions for MILAN in the conv5\_3 layer. Table \ref{tab:random noise BERT score} shows the mean and standard deviation of the F1-BERT score between the clean and manipulated descriptions of MILAN in the layers of VGG16. We observe a decrease in the F1-BERT score with increasing noise standard deviation in the higher layers suggesting that neurons are more prone to being manipulated in the higher layers.

Fig \ref{fig:Gaussian Noise/Network Dissection/Effect of random noise on concept interpretability in layer conv5_3} visualize the percentage of neurons manipulated with the gradual addition of random noise for standard deviation 0.02 and 0.05. We observe that the number of neurons manipulated increases with an increasing standard deviation of random noise. A standard deviation of 0.05 with 60\% data poisoning leads to manipulation of around 10\% neurons for Network Dissection and 25\% neurons for MILAN. This result is significant since noise addition is low cost and could happen naturally through noisy transmission, or probing dataset can be poisoned during exchange over the network.

\subsection{Designed corruption of probing dataset}
\label{Experiment-Adversarial attack}
In this section, we try to answer the following question: \textit{Can our designed corruption of the probing dataset manipulate the neuron explanations?} We visualize the results of our data corruption on Network Dissection and MILAN in Section \ref{Qualitative analysis}, perform large-scale robustness analysis for NEMs with corrupted probing dataset in Sec \ref{Large scale robustness analysis}, understand the susceptibility of categories in the Broden dataset to probing dataset corruption in section \ref{Effect on attack on categories}, and study the effect of probing dataset corruption on the model's accuracy in Sec \ref{effect of attack on model accuracy}.
\subsubsection{Qualitative analysis}
\label{Qualitative analysis}

Fig \ref{fig:Visualize attack Unit 191 MILAN} (left) visualizes the data poisoning on Network Dissection to manipulate the concept of Unit 191 in layer conv5\_3 from \textit{Bus} to \textit{Bed}. We poison a total of 2532 images out of 63296 images in the Broden dataset, which account for less than 5\% of the total images. We observe that the concept gradually changes from \textit{Bus} to \textit{Bed}, first reducing the IOU of source concept \textit{Bus} to $0.10$ for $\epsilon=2/255$, and then changing concept to Bed with a high IOU of 0.17 for $\epsilon = 3/255$.
\begin{table*}[t]
  \centering
  \scalebox{0.95}{
  \begin{tabular}{p{.2\linewidth} | >{\raggedright\let\newline\\\arraybackslash\hspace{0pt}}p{.32\linewidth} | >{\raggedright\let\newline\\\arraybackslash\hspace{0pt}}p{.32\linewidth} | p{.08\linewidth}}
    \toprule
    Layer-Unit & Clean description & \textcolor{red}{Manipulated} description & Score\\
    \hline
    \textbf{VGG16-Places365} & & & \\
    conv3\_3-42 & ``top of a building" & \textcolor{red}{``red and white colored objects"} & 0.38 \\ 
    conv4\_3-29 & ``a table" & \textcolor{red}{``body parts"} & 0.45 \\
    conv5\_3-46 & ``plants" & \textcolor{red}{``green grass"} & 0.49  \\ \hline
    \textbf{Resnet50-Imagenet} & & & \\
    layer2-6 & ``edges of objects" & \textcolor{red}{``the color red"} & 0.45\\
    layer3-29 & ``birds" & \textcolor{red}{``human faces"} & 0.27 \\
    layer4-36 & ``items with straight features" & \textcolor{red}{``rounded edges in pictures"} & 0.58 \\
    \bottomrule
  \end{tabular}}
  \caption{Sampled F1-BERT scores with untargeted data poisoning of MILAN for VGG16-Places365 and Resnet50-Imagenet. It shows the neuron explanations with score below of our selected threshold of 0.642. }
  \label{tab:Randomly sampled BERT scores from MILAN descriptions for Resnet50-Imagenet}
\end{table*}

On the other hand, Fig \ref{fig:Visualize attack Unit 191 MILAN} (right) visualizes the data poisoning on MILAN to manipulate the description of Unit 191 in layer conv5\_3 from \textit{Buses} to \textit{Beds}. We poison 800 out of 36500 images in the Places365 validation dataset, accounting for less than $3\%$ of the total images. Similar to Network Dissection, we observe that the concept gradually changes to \textit{Beds} for $\epsilon=3/255$. Further, $\epsilon=2/255$ reduces the activated area in the mask for \textit{Bus}, suggesting that our objective function is indeed reducing the activation of the source concept. Table \ref{tab:Randomly sampled BERT scores from MILAN descriptions for Resnet50-Imagenet} shows the clean and manipulated descriptions of MILAN on a few sampled neurons. Our objective function is able to manipulate descriptions to very different concepts, for example, \textit{a table} to \textit{body parts}.

\subsubsection{Large scale robustness analysis}
\label{Large scale robustness analysis}
Here, we perform a large-scale quantitative analysis of the robustness by performing targeted and untargeted corruption of probing dataset on the layers of VGG16-Places365 and Resnet50-Imagenet.

For Network Dissection, we consider an untargeted data corruption successful if the unit has been manipulated following our earlier definition, and consider a targeted data corruption successful if the manipulated concept matches the source concept. We poison less than 10\% of $\mathcal{D}_{probe}$ since the number of probing images for a concept is less than 5\% of dataset size for most concepts. Table \ref{tab:units modified by attack on Network Dissection for VGG16-Places365} shows the results of untargeted and targeted data corruption on all layers of \textit{VGG16-Places365} and \textit{Resnet50-Imagenet}  for PGD $\epsilon = 2/255, 4/255, 6/255$. Our objective function can successfully change the concepts of more than 80\% neurons in the higher layers with PGD $\epsilon = 4/255$ in the untargeted setting. We can change the concept of a neuron to a desired target concept with a high success rate of more than 50\% in higher layers with $\epsilon=6/255$. Further, we are also able to manipulate more than 60\% neuron concepts in the higher layers with $\epsilon = 6/255$ by obtaining $L_{c}(x)$ from the baseline runs as discussed in Appendix Sec A.2.1.

For MILAN, Table \ref{tab:Average BERT score and std of MILAN for untargeted attack on Resnet50-Imagenet} and Appendix Table A.7 (for type U2) shows the mean and standard deviation of F1-BERT score between clean and manipulated descriptions for  \textit{Resnet50-Imagenet} and \textit{VGG16-Places365}. We run our data corruption algorithm using projected gradient descent for different values of  $\epsilon = \frac{2}{255},\frac{4}{255},\frac{6}{255}$. We observe a consistent decrease in the F1-BERT score with the mean score being less than 0.5 for a small perturbation of $\epsilon = \frac{4}{255}$ in the higher layers of the network. Further, we poison 800 images out of 36500 images in the $\mathcal{D}_{probe}$, which constitutes less than 3\% of total images.

\subsubsection{Effect of data corruption on Broden categories}
\label{Effect on attack on categories}
\begin{figure}[h]
  \centering
  \includegraphics[width=\linewidth]{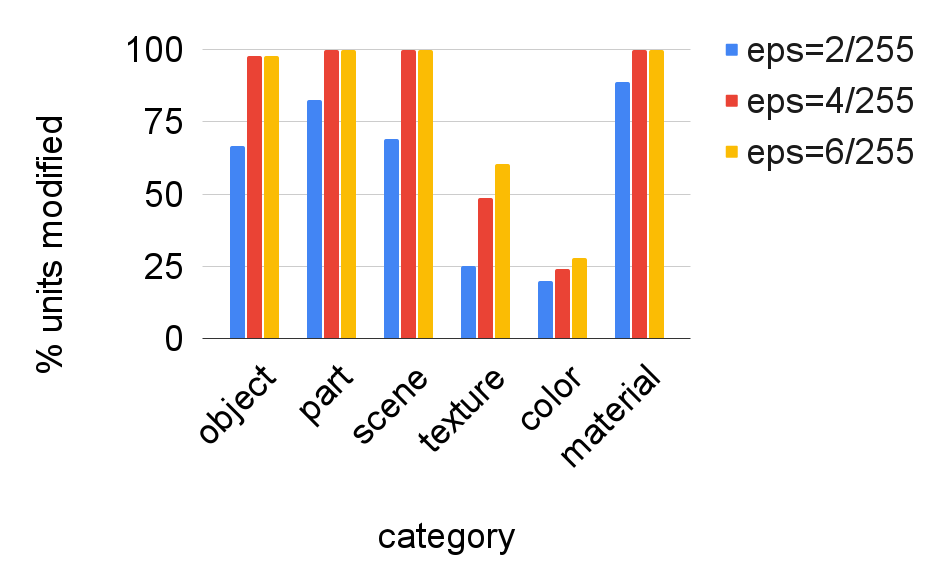}
  \caption{Percentage of concepts manipulated by category in Broden dataset with untargeted data corruption on VGG16-Places365. Higher level concepts \textit{object}, \textit{part}, \textit{scene}, and \textit{material} are more susceptible to data corruption than lower level concepts \textit{color} or \textit{texture}}
  \label{fig:concepts modified due to untargeted attack on VGG16}
\end{figure}

Broden dataset used in Network Dissection consists of images with pixel-level segmentation of concepts from six categories, which allows us to study if certain categories are more vulnerable to data corruption. We follow the experimental setting discussed in Sec \ref{Large scale robustness analysis} for the untargeted data corruption on Network Dissection, and Fig \ref{fig:concepts modified due to untargeted attack on VGG16} shows the percentage of concepts manipulated in different categories with untargeted corruption of probing dataset. Higher level concepts \textit{objects}, \textit{part}, \textit{scene}, and \textit{material} can be manipulated with a success rate of 100\% by our objective function in the untargeted setting. In contrast, lower-level concepts \textit{texture} and \textit{color} are more robust to corruption.
\subsubsection{Effect of data corruption on model accuracy}
\label{effect of attack on model accuracy}
\begin{figure}[h]
    \centering
    \includegraphics[width=\linewidth]{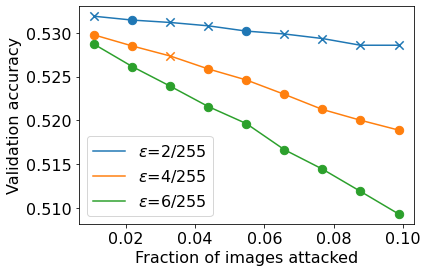}
    \caption{The effect of changing the number of poisoned images (x-axis) on the probing dataset accuracy (y-axis) for Unit 191 conv5\_3 VGG16-Places365 in MILAN. Successfully manipulated concepts are marked as circles, while unsuccessful manipulations are crossed. Our data corruption algorithm has a minimal effect on classification accuracy, indicating that our corruption can be harder to detect.}
    \label{fig:attack_vs_acc}
\end{figure}

MILAN uses the validation dataset as the probing dataset, which raises the question of using the accuracy on the validation dataset to detect data corruptions. We plot the validation accuracy of VGG16-Places365 with an increasing number of corrupted images for Unit 191-Conv 5\_3 to manipulate the explanation from the concept \textit{bus} to the concept \textit{bed}. We consider a data corruption successful if the manipulated MILAN description was semantically similar to \textit{bed}, i.e., \textit{Beds} or \textit{Furniture}. The results are visualized in Fig \ref{fig:attack_vs_acc}. We see that our data corruption algorithm can succeed with minimal effect on the validation accuracy of the model, indicating that the data corruption would be harder to detect, and even the largest perturbation only reduce model accuracy by less than 2\%.

\subsection{Comparing random and designed corruption of probing dataset}
\label{Comparing adversarial and non-adversarial attacks}
\begin{figure}
    \centering
    \includegraphics[width=\linewidth]{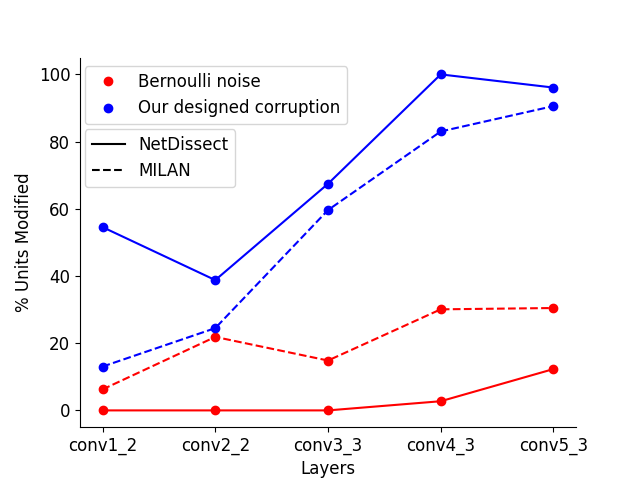}
    \caption{Comparing Bernoulli random noise and our data corruption algorithm in \textit{VGG16-Places365} with noise level and corruption magnitude $\epsilon$ set to $\frac{6}{255}$ for Network Dissection and MILAN. Higher percentage of units manipulated means that corruption technique has stronger effects. Our objective function achieves significantly higher success rate compared to Bernoulli random noise.}
    \label{fig:Comparing adversarial and non adversarial attacks}
\end{figure}

In this section, we compare random and designed data corruption by keeping noise level and corruption magnitude $\epsilon$ set to $\frac{6}{255}$. Bernoulli random noise achieves higher neuron manipulation success rate compared to Gaussian random noise and uniform random noise as shown in Appendix Sec A.1.2. Following this result, we use Bernoulli noise for random data corruption and Eq \ref{Attack: Targeted with minimum distance} for designed data corruption. The results are shown in Fig \ref{fig:Comparing adversarial and non adversarial attacks}. We observe that our data corruption algorithms achieves a significantly higher manipulation success rate than addition of random noise to the images in the probing dataset.

\section{Conclusion}
In this work, we made a significant effort to show that Neuron Explanation methods are not robust to corruptions including random noise and well-designed perturbations added to the probing dataset. Our experiments indicate that our objective function can manipulate assigned neuron concepts in both targeted and untargeted settings. Further, we show that it is easier to manipulate units in the top layer of a network that have higher-level concepts assigned to them.

\section*{Acknowledgement}
The authors would like to thank the anonymous reviewers for their valuable feedback. The authors also thank the computing resource supported in part by National Science Foundation awards CNS-1730158, ACI-1540112, ACI-1541349, OAC-1826967, OAC-2112167, CNS-2100237, CNS-2120019, the University of California Office of the President, and the University of California San Diego's California Institute for Telecommunications and Information Technology/Qualcomm Institute with special thanks to CENIC for the 100Gbps networks. T. Oikarinen and T.-W. Weng are supported by National Science Foundation under grant no. 2107189. 

\clearpage
{\small
\bibliographystyle{ieee_fullname}
\bibliography{egbib}
}

\newpage
\appendix
\counterwithin{figure}{section}
\counterwithin{table}{section}
\setcounter{table}{0}
\setcounter{figure}{0}
\onecolumn
\section{Appendix}

\subsection{Noise corruption of probing dataset}
This section extends our experiments with random noise data corruption to manipulate neuron explanations. In Section \ref{Effect of  increasing fraction of attacked images}, we will study the effect of increasing the fraction of images poisoned with Gaussian random noise for an larger set of standard deviations compared to Fig 2.b in the main text. In Sec \ref{Robustness analysis with bounded noise}, we will analyze the effect of uniform and Bernoulli bounded noise data corruption on neuron explanations.

\subsubsection{Gradual data poisoning with Gaussian noise}
\label{Effect of increasing fraction of attacked images}
Table \ref{tab:Effect of increasing fraction of attacked images - full} shows the percentage of neurons manipulated by gradual poisoning of the probing dataset with Gaussian random noise. We observe that even adding noise to $20\%$ of probing dataset can manipulate around $5\%$ neurons and $17\%$ neurons in the conv5\_3 layer for Network Dissection and MILAN respectively.
\begin{table}
  \centering
  \begin{tabular*}{\linewidth}{l@{\extracolsep{\fill}}  r@{\extracolsep{\fill}} r@{\extracolsep{\fill}} r@{\extracolsep{\fill}} r@{\extracolsep{\fill}} r@{\extracolsep{\fill}} r@{\extracolsep{\fill}} r@{\extracolsep{\fill}} r@{\extracolsep{\fill}} r@{\extracolsep{\fill}} r@{\extracolsep{\fill}}}
  \toprule
    Poisoned & \multicolumn{5}{c}{Network Dissection} & \multicolumn{5}{c}{MILAN}\\
           \cmidrule(lr){2-6} \cmidrule(lr){7-11}
    images \%   & std = 0.01 & 0.02 & 0.03 & 0.04 & 0.05 & 0.01 & 0.02 & 0.03 & 0.04 & 0.05 \\ \midrule
    20  & 0.586 & 1.367 & 3.125 & 3.906 & 4.492 & 6.45 & 11.33 & 13.45 & 12.5 & 16.21    \\
    40  & 0.195 & 2.930 & 4.492 & 5.859 & 6.836  & 8.79 & 13.48 & 17.97 & 19.73 & 21.48  \\
    60  & 0.781 & 3.711 & 6.641 & 9.375 & 11.328  & 9.38 & 16.21 & 20.31 & 23.04 & 24.80 \\
    80  & 0.781 & 4.883 & 8.398 & 12.109 & 15.430 & 11.72 & 18.55 & 24.41 & 25.58 & 27.34 \\
    100 & 0.781 & 6.055 & 10.352 & 13.477 & 17.578 & 18.36 & 28.71 & 35.94 & 38.28 & 41.80 \\
    \bottomrule
    \end{tabular*}
  \caption{Percentage of units manipulated (higher
means corruption technique has stronger effects) in \textit{VGG16-Places365} conv5\_3 by gradual poisoning of the probing dataset with Gaussian random noise for Network Dissection.}
  \label{tab:Effect of increasing fraction of attacked images - full}
\end{table}

\subsubsection{Robustness analysis with bounded noise}
\label{Robustness analysis with bounded noise}
\begin{figure}[b]
    \centering
    \begin{subfigure}{\linewidth}
      \centering
      \includegraphics[width=\linewidth]{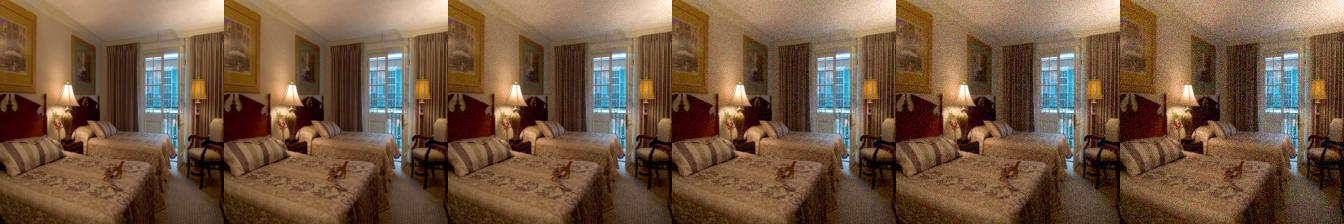}
      \caption{Gaussian noise}
      \label{}
    \end{subfigure}
    \begin{subfigure}{\linewidth}
      \centering
      \includegraphics[width=\linewidth]{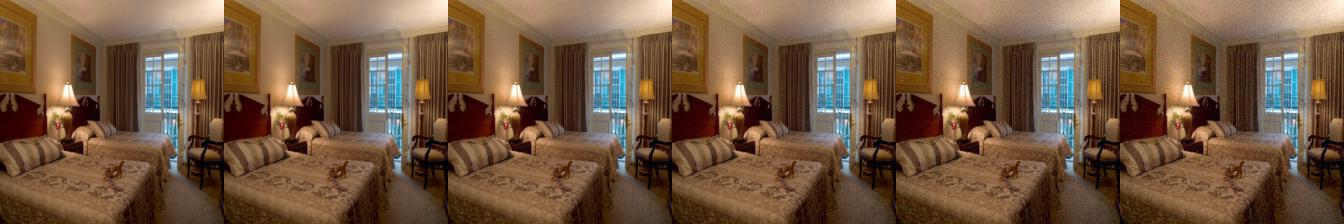}
      \caption{Uniform noise}
      \label{}
    \end{subfigure}
    \begin{subfigure}{\linewidth}
      \centering
      \includegraphics[width=\linewidth]{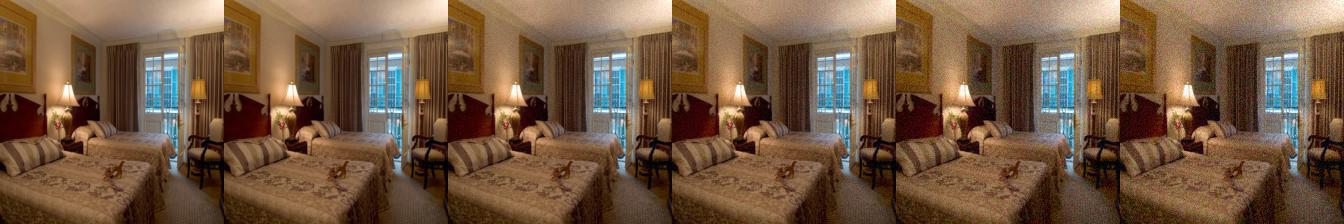}
      \caption{Bernoulli noise}
      \label{}
    \end{subfigure}
\caption{Visualizing images poisoned with (a) Gaussian, (b) Uniform, and (c) Bernoulli noise. The first image of each row visualizes the original image, followed poisoned images with a standard deviation from $0.01$ to $0.05$ in step size of $0.01$. The images are visually unchanged even with noise std=0.05 added to it.}
\label{fig:visualize images with bounded noise}
\end{figure}

In this section, we study the effect of data corruption with bounded noise on neuron explanations. We poison probing dataset with uniform and Bernoulli random noise and obtain robustness estimates for \textit{VGG16-Places365}. Fig \ref{fig:visualize images with bounded noise} visualizes images in the probing dataset with added Gaussian, uniform, and Bernoulli noise. Even with noise standard deviation of 0.05, the images are visually unchanged, making it hard to discern data corruption through manual examination.
\begin{table}[!h]
  \centering
  \begin{tabular*}{\linewidth}{l@{\extracolsep{\fill}}  r@{\extracolsep{\fill}} r@{\extracolsep{\fill}} r@{\extracolsep{\fill}} r@{\extracolsep{\fill}} r@{\extracolsep{\fill}} r@{\extracolsep{\fill}}}
  \toprule
    Layer & Type & std = 0.01 & 0.02 & 0.03 & 0.04 & 0.05\\ \midrule
    \multirow{3}{*}{conv3\_3} & G & 0.0 & 0.0 & 0.0 & 0.390 & 0.781 \\ 
    & U & 0.0 & 0.0 & 0.0 & 0.0 & 0.0 \\
    & B & 0.0 & 0.0 & 0.0 & \textbf{1.172} & \textbf{1.953} \\ \midrule
    \multirow{3}{*}{conv4\_3} & G & 0.195 & 1.171 & 3.125 & 4.687 & 6.0546 \\
    & U & 0.195 & 0.390 & 1.953 & 3.320 & 5.860 \\
    & B & 0.195 & \textbf{2.539} & \textbf{5.664
    }& \textbf{10.938} & \textbf{12.500} \\ \midrule
    \multirow{3}{*}{conv5\_3} & G & 0.781 & 5.859 & 10.156 & 13.281 & 17.773 \\
    & U & \textbf{0.586} & 1.171 & 7.6171 & 13.867 & 21.679 \\
    & B & 0.390 & \textbf{12.500} & \textbf{21.680} & \textbf{36.328} & \textbf{41.601} \\ \midrule
    \end{tabular*}
  \caption{Percentage of neurons manipulated  (higher means corruption technique has stronger effects) in \textit{VGG16-Places365} by addition of Gaussian noise(G), Uniform noise(U), and Bernoulli noise(B) for Network Dissection. We do not observe any successful manipulation in the layers conv1\_1 and conv2\_2. \textbf{Highlighted} values indicate maximum percentage of units manipulated for a given layer and noise level. Bernoulli noise leads to highest number of manipulated neurons.} 
  \label{tab:units modified with bounded noise with network dissection}
\end{table}
\begin{table*}
  \centering
  \begin{tabular*}{\linewidth}{l@{\extracolsep{\fill}}  r@{\extracolsep{\fill}} r@{\extracolsep{\fill}} r@{\extracolsep{\fill}} r@{\extracolsep{\fill}} r@{\extracolsep{\fill}} r@{\extracolsep{\fill}}}
  \toprule
    & & \multicolumn{5}{c}{Noise}\\
           \cmidrule(lr){3-7}
    Layer & Type & 0.01 & 0.02 & 0.03 & 0.04 & 0.05\\
    \midrule
    \multirow{3}{*}{conv1\_2} & G & $0.936 \pm 0.172$ & \textbf{0.892 $\pm$ 0.208} & $0.884 \pm 0.222$ & $0.870 \pm 0.196$ & \textbf{0.843 $\pm$ 0.203} \\
    & U & $0.967 \pm 0.117$ & $0.933 \pm 0.152$ & $0.900 \pm 0.202$ & $0.865 \pm 0.230$ & $0.876 \pm 0.192$ \\
    & B & \textbf{0.918 $\pm$ 0.189} & $0.896 \pm 0.189$ & \textbf{0.867 $\pm$ 0.206} & \textbf{0.843 $\pm$ 0.202} & $0.866 \pm 0.213$ \\
    \midrule
    \multirow{3}{*}{conv2\_2} & G & $0.918 \pm 0.169$ & $0.862 \pm 0.202$ & $0.856 \pm 0.212$ & $0.829 \pm 0.214$ & $0.848 \pm 0.206$ \\
    & U & $0.933 \pm 0.155$ & $0.910 \pm 0.177$ & $0.867 \pm 0.208$ & \textbf{0.853 $\pm$ 0.216} & $0.866 \pm 0.200$ \\
    & B & \textbf{0.914 $\pm$ 0.176} & \textbf{0.870 $\pm$ 0.201} & \textbf{0.849 $\pm$ 0.205} & $0.860 \pm 0.204$ & \textbf{0.836 $\pm$ 0.209} \\
    \midrule
    \multirow{3}{*}{conv3\_3} & G & $0.922 \pm 0.181$ & $0.886 \pm 0.214$ & $0.845 \pm 0.226$ & $0.813 \pm 0.235$ & $0.813 \pm 0.238$ \\
    & U & $0.937 \pm 0.167$ & $0.909 \pm 0.192$ & $0.889 \pm 0.201$ & $0.875 \pm 0.213$ & $0.842 \pm 0.224$ \\
    & B & \textbf{0.928 $\pm$ 0.167} & \textbf{0.852 $\pm$ 0.232} & \textbf{0.836 $\pm$ 0.222} & \textbf{0.793 $\pm$ 0.247} & \textbf{0.795 $\pm$ 0.252} \\
    \midrule
    \multirow{3}{*}{conv4\_3} & G & $0.896 \pm 0.207$ & $0.834 \pm 0.238$ & $0.778 \pm 0.259$ & $0.765 \pm 0.259$ & $0.735 \pm 0.263$ \\
    & U & $0.931 \pm 0.174$ & $0.881 \pm 0.218$ & $0.840 \pm 0.238$ & $0.820 \pm 0.243$ & $0.785 \pm 0.259$ \\
    & B & \textbf{0.897 $\pm$ 0.208} & \textbf{0.812 $\pm$ 0.249} & \textbf{0.797 $\pm$ 0.248} & \textbf{0.746 $\pm$ 0.264} & \textbf{0.733 $\pm$ 0.270} \\
    \midrule
    \multirow{3}{*}{conv5\_3} & G & $0.886 \pm 0.223$ & $0.832 \pm 0.251$ & $0.787 \pm 0.269$ & $0.767 \pm 0.269$ & $0.742 \pm 0.279$ \\
    & U & $0.935 \pm 0.174$ & $0.868 \pm 0.237$ & $0.842 \pm 0.253$ & $0.807 \pm 0.261$ & $0.801 \pm 0.257$ \\
    & B & \textbf{0.888 $\pm$ 0.224} & \textbf{0.823 $\pm$ 0.256} & \textbf{0.783 $\pm$ 0.275} & \textbf{0.758 $\pm$ 0.271} & \textbf{0.730 $\pm$ 0.283} \\
    \bottomrule
    \end{tabular*}
  \caption{Average F1-BERT score (lower score
means our corruption technique has stronger effects) between clean and manipulated descriptions in \textit{VGG16-Places365} with Gaussian noise(G), Uniform noise(U), and Bernoulli noise(B) data corruption for MILAN. Lower score indicates higher dissimilarity and more successful untargeted data corruption. \textbf{Highlighted} values indicate minimum F1-BERT score for a given layer and noise level. Bernoulli noise leads to lowest F1-BERT score.}
  \label{tab:units modified with bounded noise with MILAN}
\end{table*}
\begin{table*}[t]
  \begin{tabular}{p{.3\linewidth} | >{\raggedright\let\newline\\\arraybackslash\hspace{0pt}}p{.32\linewidth} | >{\raggedright\let\newline\\\arraybackslash\hspace{0pt}}p{.32\linewidth} | p{.08\linewidth}}
    \toprule
    Clean description & \textcolor{red}{Manipulated} description & Score\\
    \hline
    ``vehicle windows'' & \textcolor{red}{``car windows"} & 0.942 \\
    ``doors'' & \textcolor{red}{``wall"} & 0.858 \\
    ``the top of round objects" & \textcolor{red}{``circular shaped objects"} & 0.659 \\ 
    ``containers" & \textcolor{red}{``circular object"} & 0.638 \\
    ``doors" & \textcolor{red}{``buildings"} & 0.604  \\ 
    ``signs and grids'' & \textcolor{red}{``red colored object with text"} & 0.564 \\
    ``blue areas in pictures'' & \textcolor{red}{``blue skies"} & 0.552 \\
    ``trees and flowers'' & \textcolor{red}{``green colored objects"} & 0.422 \\
    ``vertical lines'' & \textcolor{red}{``fencing"} & 0.327 \\\hline
  \end{tabular}
  \caption{Sampled F1-BERT scores with untargeted data poisoning of MILAN}
  \label{Sampled F1-BERT scores with untargeted data poisoning of MILAN}
\end{table*}

\paragraph{Network Dissection} Table \ref{tab:units modified with bounded noise with network dissection} shows the percentage of neurons manipulated in \textit{VGG16-Places365} with bounded noise data corruption. Bernoulli noise manipulates the highest percentage of neurons at different noise levels, with a maximum of $42\%$ units in the conv5\_3 layer.

\paragraph{MILAN} Table \ref{tab:units modified with bounded noise with MILAN} shows the average F1-BERT score between clean and manipulated descriptions with bounded noise data corruption. Similar to Network Dissection, we observe that Bernoulli noise results in maximum change in the neuron descriptions, with a noise level of 0.05 reducing the average F1-BERT score to 0.73 in the conv5\_3 layer.
\begin{figure}
    \centering
    \includegraphics[width=.65\linewidth]{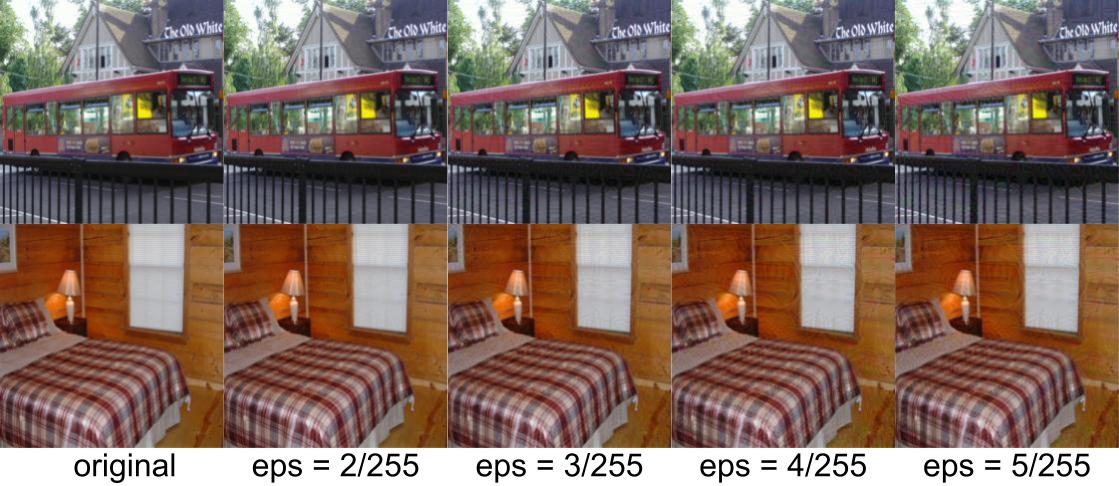}
    \caption{Visualizing images poisoned with designed data corruption for different values of corruption magnitude  $\epsilon$.}
    \label{fig:Visualize attacked Broden images}
\end{figure}

\subsection{Designed corruption of probing dataset}
\begin{figure}[h]
\centering
\includegraphics[width=\linewidth]{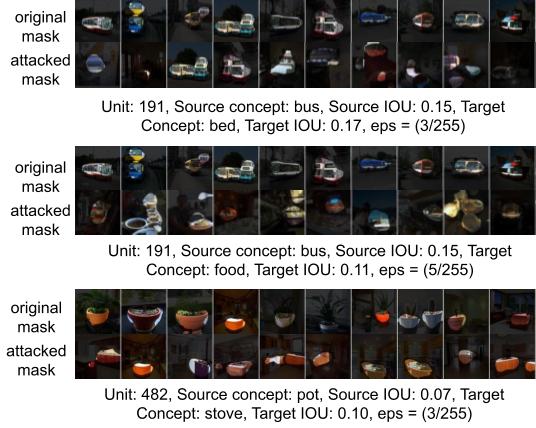}
\caption{Successful targeted data corruption with least corruption magnitude $\epsilon$ to manipulate unit 191 and unit 482. First row in subfigure visualizes top activating images from $\mathcal{D}_{probe}$. Second row in each subfigure highlights pixels with value greater than activation threshold $T_{neuron}$.}
\label{fig:Targeted attack on VGG16 conv5_3}
\end{figure}

This section extends our experiments with the designed data corruption to manipulate neuron explanations. We present our results with untargeted data corruption without using ground-truth segmentation for Network Dissection in Sec \ref{Untargeted data corruption on Network Dissection with $L_{c}(x)$ from general NEMs}, discuss challenges with targeted data corruption for MILAN in Sec \ref{Targeted Attack on MILAN}, provide an ablation study with a reduced form of Eq 9 in untargeted setting in Sec \ref{Ablation: Untargeted Attack}, and perform robustness analysis of adversarially-trained Resnet50 model in Sec \ref{Robustness of Adversarially-trained models}. Fig \ref{fig:Visualize attacked Broden images} visualizes poisoned images with our objective function, and Fig \ref{fig:Targeted attack on VGG16 conv5_3} visualizes successful targeted data corruption for Network Dissection on selected neuron units. 

\subsubsection{Untargeted data corruption on Network Dissection with $L_{c}(x)$ from uncorrupted runs}
\label{Untargeted data corruption on Network Dissection with $L_{c}(x)$ from general NEMs}
In this section, we analyze untargeted data corruption for Network Dissection under the assumption that the ground-truth segmentation information is not accessible, and hence we cannot obtain $L_{c}(x)$ directly for data corruption with Eq 9. This is a typical setting when Network Dissection (and other NEMs) are offered as a cloud service, with the user providing the probing dataset and the per-pixel segmentation generation being handled securely in the cloud. We follow our formulation in Sec 3.2 for obtaining $L_{c}(x)$ from the general NEMs. The results are shown in Table \ref{tab:Network Dissection from baseline runs}. We observe that the designed data corruption is less effective than using the baseline segmentation data; however, we can achieve $45\%$ success rate on $conv5\_3$ layer by poisoning less than $10\%$ images. This shows that hiding the $\texttt{sim}$ function is not a viable defense against our designed data corruption method.
\begin{table}[!h]
\centering
\scalebox{.9}{\begin{tabular*}{\linewidth}{l@{\extracolsep{\fill}}  r@{\extracolsep{\fill}} r@{\extracolsep{\fill}} r@{\extracolsep{\fill}}}
\toprule
Layer & $\epsilon=\frac{2}{255}$ & $\frac{4}{255}$ &  $\frac{6}{255}$\\
\midrule
conv4\_3 & 5.41 & 8.11 & 13.51\\
\midrule
conv5\_3 & 5.13 & 23.08 & 46.15\\
\bottomrule
\end{tabular*}}
\caption{Percentage of units manipulated (higher
means our corruption technique has stronger effects) in \textit{VGG16-Places365} by untargeted data corruption with $L_{c}(x)$ obtained from uncorrupted runs for Network Dissection. We can obtain 45\% manipulation success rate without using ground truth segmentation data.}
\label{tab:Network Dissection from baseline runs}
\end{table}

\subsubsection{Targeted data corruption on MILAN}
\label{Targeted Attack on MILAN}
\begin{figure}[!h]
    \centering
    \includegraphics[width=.35\linewidth]{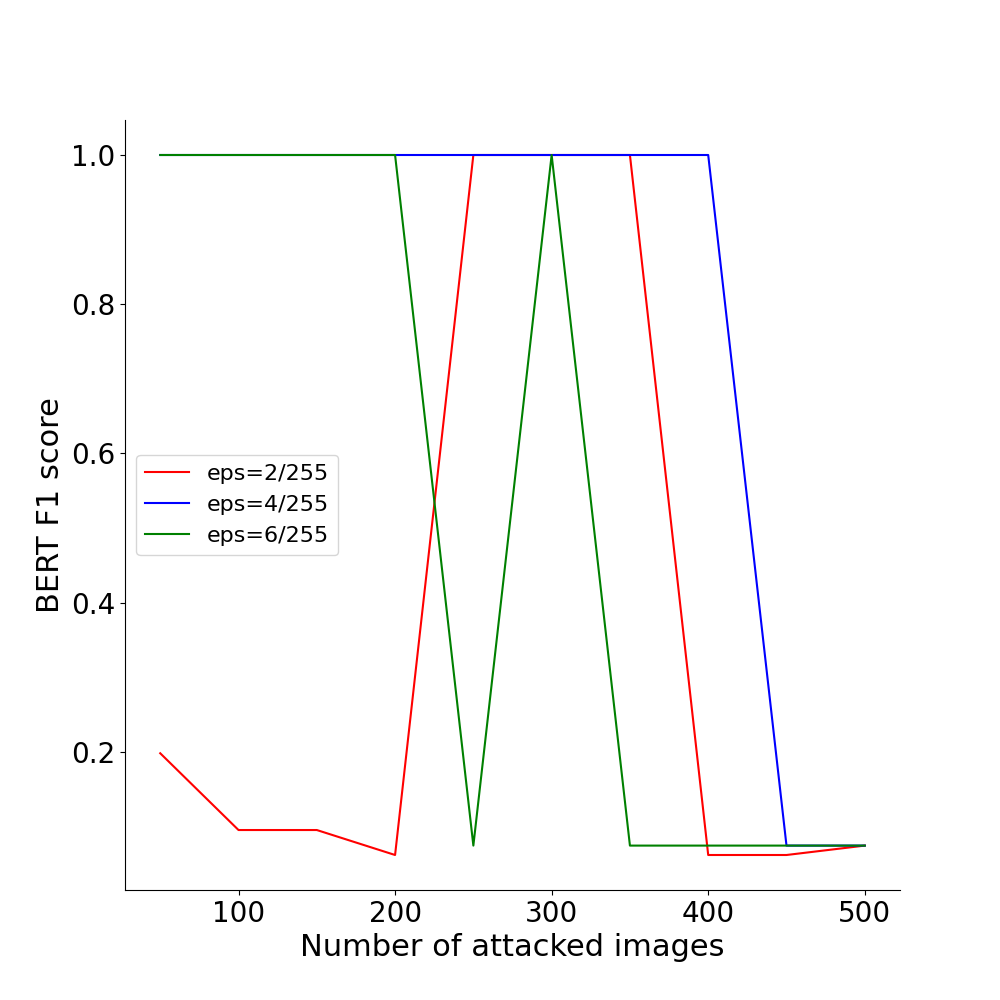}
    \caption{F1-Bert score between clean and manipulated description with increasing number of poisoned images for Unit 191 conv5\_3 of \textit{VGG16-Places365} in MILAN. Higher score means manipulated description and target description are similar, with F1-BERT score being 1.0 if they match. The F1-BERT score increases and then decreases, implying that the manipulated description is a function of the number of poisoned images.}
    \label{fig:Changing number of attacked images MILAN}
\end{figure}

The targeted data corruption aims to manipulate clean descriptions of MILAN to a target description by poisoning images with designed perturbations. We consider a targeted data corruption successful if the F1-BERT score between manipulated and target description is more significant than $0.642$. We observe in our experiments that the targeted designed data corruption on MILAN can manipulate a maximum of $40\%$ neurons in \textit{VGG16-Places365}. This can be explained by the fact that the manipulated description is dependent on the number of poisoned images. MILAN obtains $L_{c}(x)$ for computing corruption using standard (non-corrupted) runs and requires hyperparameter for the number of poisoned images. We argue that the manipulated description is dependent on the number of poisoned images, and increasing the number of poisoned images after a threshold results affects the activations of unrelated concepts. Fig \ref{fig:Changing number of attacked images MILAN} shows the targeted designed data corruption on Unit 191 to change its concept from ``buses" to ``beds" with the varying number of poisoned images. We observe that the F1-BERT score between manipulated and target description decreases to a very low value if the number of corrupted images exceeds 400, implying that the manipulated description is very different from target description.

\subsubsection{Ablation: Untargeted data corruption}
\label{Ablation: Untargeted Attack}
An alternative untargeted data corruption objective for neuron $i$ with concept $c_{i}^{*}$, and image $x_{j} \in \mathcal{D}_{probe}$ can be formulated as
\begin{equation} \label{Attack: Untargeted with minimum distance}
    \begin{aligned}
        & \underset{\delta_{j}}{\min} &&  act_{i,c_{i}^{*}}^{avg}(x_{j}+\delta_{j})  \\
        & \textrm{s.t.} && ||\delta_{j}||_{\infty} \leq \epsilon \\
        &&& x_{j}+\delta_{j} \in [0,1]^l\\
    \end{aligned}
\end{equation}
This formulation can be intuitively understood as trying to reduce the activations of pixels associated with the source category. We refer to this formulation as U1 and our formulation in Eq 9 as U2. Table \ref{tab:ablation:units modified by attack on Network Dissection for VGG16-Places365} and Table \ref{tab:ablation:Average BERT score and std of MILAN for untargeted attack on MILAN} compare the strength of untargeted data corruption with U1 and U2 for Network Dissection and MILAN respectively. We observe that U2 consistently outperforms U1, implying that using a random target label results in a stronger data corruption in an untargeted setting.
\begin{table}[t]
  \centering
  \scalebox{0.9}{\begin{tabular*}{\linewidth}{l@{\extracolsep{\fill}} r@{\extracolsep{\fill}} r@{\extracolsep{\fill}} r@{\extracolsep{\fill}}
  r@{\extracolsep{\fill}}}
  \toprule
  Layer & type & $\epsilon=\frac{2}{255}$ & $\frac{4}{255}$ &  $\frac{6}{255}$ \\
  \midrule
\multirow{2}{*}{conv1\_2} & U1 & 0.0 & 0.0 & 0.0 \\
                          & U2 & \textbf{47.73} & \textbf{52.27} & \textbf{54.55} \\
  \midrule
\multirow{2}{*}{conv2\_2} & U1 & 0.0 & 0.0 & 0.0 \\
                          & U2 & \textbf{4.08} & \textbf{22.45} & \textbf{38.78}  \\
  \midrule
\multirow{2}{*}{conv3\_3} & U1 & 6.67 & 33.33 & 46.67 \\
                          & U2 & \textbf{34.88} & \textbf{58.14} & \textbf{67.44} \\
  \midrule
\multirow{2}{*}{conv4\_3} & U1 & 53.85 & 53.85 & 69.23 \\
                          & U2 & \textbf{57.45} & \textbf{95.74} & \textbf{100.0} \\
  \midrule
\multirow{2}{*}{conv5\_3} & U1 & 38.46 & 53.85 & 53.85 \\
                          & U2 & \textbf{70.59} & \textbf{96.08} & \textbf{96.08} \\
  \bottomrule
  \end{tabular*}}
  \caption{Percentage of units manipulated (higher
score means corruption technique has stronger effects) in \textit{VGG16-Places365} by untargeted data corruption with objective function U1 and U2 for Network Dissection. \textbf{Highlighted} values indicates stronger corruption between U1 and U2 for a given layer and corruption magnitude $\epsilon$. Objective function U2 consistently outperforms objective function U1 and achieves a higher manipulation success rate.}
  \label{tab:ablation:units modified by attack on Network Dissection for VGG16-Places365}
\end{table}
\begin{table}[!h]
  \centering
  \scalebox{0.9}{\begin{tabular*}{\linewidth}{l@{\extracolsep{\fill}} r@{\extracolsep{\fill}} r@{\extracolsep{\fill}} r@{\extracolsep{\fill}}  r@{\extracolsep{\fill}}
r@{\extracolsep{\fill}} }
  \toprule
Layer & type & $\epsilon = \frac{2}{255}$ & $\epsilon = \frac{4}{255}$ &  $\epsilon = \frac{6}{255}$ \\
\midrule
\multirow{2}{*}{conv1\_2} & U1 & 0.91 $\pm$ 0.23 &  0.95 $\pm$ 0.17 & 0.92 $\pm$ 0.20 \\
                          & U2 & \textbf{0.89 $\pm$ 0.26} &  \textbf{0.89 $\pm$ 0.26} & \textbf{0.86 $\pm$ 0.26} \\
\midrule
\multirow{2}{*}{conv2\_2} & U1 & 0.95 $\pm$ 0.13 & 0.89 $\pm$ 0.18 & 0.88 $\pm$ 0.19 \\
                          & U2 & \textbf{0.94 $\pm$ 0.16} & \textbf{0.90 $\pm$ 0.19} & \textbf{0.84 $\pm$ 0.21}\\
\midrule
\multirow{2}{*}{conv3\_3} & U1 & 0.86 $\pm$ 0.19 & 0.76 $\pm$ 0.24 & 0.71 $\pm$ 0.24 \\
                          & U2 & \textbf{0.79 $\pm$ 0.28} & \textbf{0.73 $\pm$ 0.26} & \textbf{0.66 $\pm$ 0.24}\\
\midrule
\multirow{2}{*}{conv4\_3} & U1 & 0.73 $\pm$ 0.25 & 0.69 $\pm$ 0.26 & 0.68 $\pm$ 0.26 \\
                          & U2 & \textbf{0.63 $\pm$ 0.27} & \textbf{0.58 $\pm$ 0.24} & \textbf{0.51 $\pm$ 0.18}\\
\midrule
\multirow{2}{*}{conv5\_3} & U1 & 0.75 $\pm$ 0.27 & 0.59 $\pm$ 0.27 & 0.53 $\pm$ 0.21 \\
                          & U2 & \textbf{0.54 $\pm$ 0.24} & \textbf{0.46 $\pm$ 0.16} & \textbf{0.47 $\pm$ 0.14} \\
\bottomrule
\end{tabular*}}
\caption{Average F1-BERT score (lower score
means our corruption technique has stronger effects) in \textit{VGG16-Places365} by untargeted data manipulation with objective function U1 and U2 for MILAN. \textbf{Highlighted} values indicates strong corruption between U1 and U2 for a given layer and  corruption magnitude $\epsilon$. F1-BERT score between similar strings has its maximum value 1.0, and lower F1-BERT score indicates higher dissimilarity and more successful untargeted data corruption. Objective function U2 consistently outperforms objective function U1 and achieves a lower F1-Bert Score.}
\label{tab:ablation:Average BERT score and std of MILAN for untargeted attack on MILAN}
\end{table}

\subsubsection{Robustness of Adversarially-trained models}
\label{Robustness of Adversarially-trained models}
In this section, we manipulate descriptions of adversarially-trained ($\epsilon$=$\frac{2}{255}$) Resnet50. The attack success rate (ASR) on Network Dissection decreases to 41.2\%, 76.0\%, 88.0\% for PGD $\epsilon$=$\frac{2}{255}$, $\frac{4}{255}$, $\frac{6}{255}$ in layer 3, from 75.3\%, 98.6\%, 98.6\% of the standard model respectively. This result suggests that adversarial training may only help to alleviate our proposed attacks a bit since the ASR is still pretty high, demonstrating the need of a stronger defense.

\end{document}